\documentclass[journal]{IEEEtran}

\usepackage{graphicx}
\graphicspath{ {./Fig/} }
\usepackage{booktabs}
\usepackage{amsmath}
\usepackage{tabularx}
\usepackage{algorithm}
\usepackage{algpseudocode}
\usepackage{subcaption}

\begin{document}

\title{RFID based Health Adherence Medicine Case Using Fair Federated Learning}

\author{Ali Kamrani khodaei, Sina Hajer Ahmadi}%

\maketitle

\begin{abstract}
Medication nonadherence significantly reduces the effectiveness of therapies, yet it remains prevalent among patients. Nonadherence has been linked to adverse outcomes, including increased risks of mortality and hospitalization. Although various methods exist to help patients track medication schedules, such as the Intelligent Drug Administration System (IDAS) and Smart Blister, these tools often face challenges that hinder their commercial viability. Building on the principles of dosage measurement and information communication in IoT, we introduce the Smart Pill Case—a smart health adherence tool that leverages RFID-based data recording and NFC-based data extraction. This system incorporates a load cell for precise dosage measurement and features an Android app to monitor medication intake, offer suggestions, and issue warnings. To enhance the effectiveness and personalization of the Smart Pill Case, we propose integrating federated learning into the system. Federated learning allows the Smart Pill Case to learn from medication adherence patterns across multiple users without compromising individual privacy. By training machine learning models on decentralized data collected from various Smart Pill Cases, the system can continuously improve its recommendations and warnings, adapting to the diverse needs and behaviors of users. This approach not only enhances the tool's ability to support medication adherence but also ensures that sensitive user data remains secure and private.
\end{abstract}


\IEEEpeerreviewmaketitle

\section{Introduction}
Federated learning is a decentralized machine learning approach where models are trained across multiple devices or servers holding local data, without sharing the data itself \cite{fedavg,kairouz2021advances,li2020review,9833972}. This technique addresses privacy concerns by ensuring that sensitive information remains on the local devices while still enabling the collaborative development of robust models. Despite its advantages, federated learning has faced challenges, particularly concerning fairness. As models are trained on data from diverse sources, there is a risk that the resulting models may perform disproportionately well for certain groups or individuals, potentially exacerbating existing biases and inequities.

Several techniques have been proposed to tackle fairness issues in federated learning (FL). The AFL method \cite{fedfa, afl} focuses on "good-intent fairness," aiming to protect the worst-case performance of any client. However, this approach is limited to small networks with dozens of clients, as it treats each client as a separate domain, potentially impacting generalizability. Li \emph{et al.} \cite{qfedavg} propose a fair resource allocation method to balance overall performance and fairness. Hu \emph{et al.} \cite{fedmgda} identify a common degradation direction applicable to all clients without compromising individual benefits and show resilience to loss inflation attacks. Additionally, Li \emph{et al.} \cite{fedmultitaskcc,10487854} investigate the trade-off between general robustness and fairness by personalizing models for each client.

Fair federated learning seeks to address these challenges by incorporating fairness considerations into the training process. This involves designing algorithms that ensure equitable performance across different demographic groups, such as age, gender, or socioeconomic status \cite{afl,fedmgda,hamidi2024adafed,10381881}. By integrating fairness constraints and metrics into the federated learning framework, researchers aim to create models that not only respect user privacy but also avoid reinforcing or amplifying biases present in the data. This approach ensures that the benefits of federated learning are distributed more evenly, promoting fairness and inclusivity in the resulting machine learning models.

Class imbalance in federated learning arises when the distribution of classes across different clients is highly uneven, which can lead to models that are biased towards the majority class and perform poorly on minority classes. In federated learning settings, this issue is compounded by the decentralized nature of the data, where each client may have a unique and unbalanced dataset. Techniques to address class imbalance in federated learning include data augmentation strategies, such as synthetic data generation or resampling techniques, which aim to balance the class distribution within each client's local dataset. Additionally, federated averaging algorithms can be adapted to weigh updates from clients differently based on their class distribution, helping to mitigate the bias towards majority classes. Advanced approaches also involve incorporating fairness constraints or regularization techniques into the training process to ensure that the global model maintains equitable performance across all classes. Addressing class imbalance effectively is crucial for developing robust federated models that can generalize well and provide fair predictions across diverse and unevenly distributed datasets \cite{fedwithnoniid,10619204,ontheconvergence}.

Incorporating fairness into federated learning requires a careful balance between privacy, model performance, and equity. Techniques such as personalized fairness constraints, differential privacy, and fairness-aware aggregation methods are being explored to address these challenges. As federated learning continues to evolve, ensuring fair outcomes will be crucial for its widespread adoption and effectiveness, particularly in applications where equitable treatment of all users is essential. By addressing these issues, fair federated learning aims to harness the benefits of decentralized data processing while upholding the principles of fairness and justice in machine learning \cite{9833972}.

Medication adherence is a critical part of the treatment process as medications are the primary tool to prevent and cure chronic illness [4]. However, it is still very common that the patients frequently do not adhere and it was estimated that half of the prescriptions were not properly followed in the USA [4]. It was discovered that medication nonadherence may result in increased mortality and the probability of hospitalization [2] as the patients who do not follow the prescription may have a 29\% higher risk of hospitalization [5]. Medication adherence is also associated with \$611 million to \$1.16 billion annual costs saving on pharmacy, outpatient and inpatient costs [5]. Therefore, a proper health adherence tool will improve the quality of care to the patients while saving a huge potential cost. Motivated by this idea, many works of literature were providing solutions for dosage measuring to evaluate patient adherence.\\
\indent Products such as the Medication Event Monitoring System (MEMS) Cap record the date and time when the patient opens the vial and the recorded event can be transferred wirelessly to a MEMS reader. The MEMS Cap is also equipped with an optional LCD screen on the top to display the number of medicine taken in the last 24 hours [6]. The Helping-Hand from Bang \& Olufsen Medicom is another intelligent drug administration system [7] that has an outer case for the specifically sized blister pack and record the time when pills were taken out from the blister pack and is able to provide a reminder to patients through visible light and audible sound [3]. Another similar product is the Smart Blister, it is based on a radio frequency identification (RFID)-enabled self-adhesive label which can be fixed on the existing blister pack, the label that affixed on the blister pack contain event-detection microcircuitry which records the time when a pill was removed from the blister pack. The patients then return the blister pack to the pharmacy where the data will be downloaded through an NFC device and stored in the database for further analysis [3]. All of these products were designed to record the metadata of the dosage and obtained later through different communication techniques for data managing and processing. Therefore, the fundamental goal of such designs should follow the similar ideas of metadata recording, data transmission and data processing.
\\
\indent The internet of things (IoT) is the concept of connecting any device in real life with the Internet for data sharing regarding the collection of desired data [8]. By combining the idea of IoT with the smart health adherence tool, a smart health adherence device named as the Smart Pill Case is proposed. It uses a load cell to measure the pill weights inside the pill container and functions as the dosage measuring technique. The pill weights will be processed and written on an NFC tag through RFID. A Smart Phone application will then read the NFC tag to extract the weight and convert it to dosage. Furthermore, the dosage information can be directly visible to the user and provide a warning on over or insufficient medication. Take the concept of IoT, the dosage data can be shared through Smart Phone Internet to a cloud database for data management and visible to whom that may concern such as a doctor or family member.

\section{Literature Review}

In this section, the three smart health adherence tools: MEMS Cap, Helping-Hand and Smart Blister will be discussed in terms of their feature, system architecture, performance test and pros/cons comparison by reviewing the literature from multiple sources.


\subsection{MEMS Cap}
\subsubsection{Feature and System Architecture}
\begin{figure}[h]
    \centering
    \includegraphics[scale=0.5]{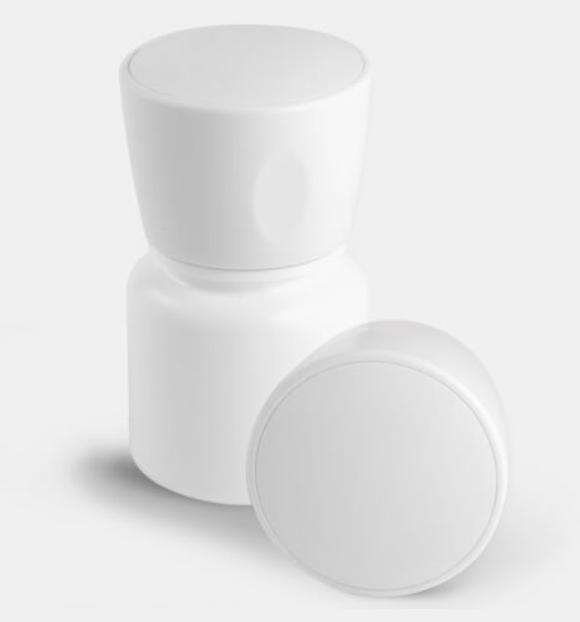} 
    \caption{MEMS Cap (the top part) will detect the date and time when the bottle is opened over 3s, the pill bottle can be any bottle as long as it fit with the cap [9].}
    \label{fig:Cap}
\end{figure}

Medical Electronic Monitoring System (MEMS) Cap from the AARDEX Group that is shown in Fig. \ref{fig:Cap} records the event of bottle opening to provide feedback regarding adherence. Every time the patient opens the vial, the processor inside the device will record the date and time of this action and it is able to store up to 4,000 dosing events, the data stored can be read through MEMS reader and analyzed with MEMS ADHERENCE SOFTWARE\cite{aardexgroup2021mems}. The data that was stored will transfer from the MEMS Cap to the Reader by wirelessly contactless communication[9]. The battery can support everyday use for up to 3 years and the LCD display on the top that display the number of medicine taken in the last 24 hours [6] is optional. It is available on the market that costs \$110 for each MEMS Cap [10].\\
\indent The overall system can be separated into three parts: the MEMS Cap, the pill bottle and the MEMS Reader. The MEMS Cap is equipped with a microelectronic circuit to detect and register the event, however, the bottle need to be open up to 3 seconds to ensure the medicine was properly taken [11]. The stored events will be stored in the micro-controller inside the MEMS Cap and wait for the MEMS Reader to extract the data. The MEMS Reader equip with the NFC reader and writer to communicate with the MEMS Cap and read the stored event data. The pill bottle can be any bottle that fits with the MEMS Cap as it does not have any extra functions other than containing the medicines.

\subsubsection{Performance Test}
There are many works of literature about the performance of MEMS Cap, and in this paper, two of them will be discussed regarding the adherence improvement in patients with schizophrenia and HAART, where medication adherence is crucial to the cure of the disease.\\
\indent For the study of Antipsychotic treatment dosing, irregular dosing occurs very often amount schizophrenia patients, the dosing profile of 74 outpatients with schizophrenia were observed for 3 months, it was found that 75.5\% patients were adherent to the treatment and 35\% patients followed the prescription perfectly while the adherent did not decrease during the 3-month period [12].\\
\indent For the study of Highly Active Antiretroviral Treatment (HAART), 265 patients were observed for a 12-month period for the measurement of adherence. As a result, more than 90\% of the HIV outpatients followed the prescribed drug dosing regimen [13].\\
\indent It was found in both literature that the willingness of adherence reaches the minimum during weekends, yet the improvement of patients' adherence was not significantly increased with the help of MEMS Cap. The potential causes may be the patients do not have access to the real-time results as the data were recorded during their visit to the clinic. Also, the warning was only delivered on the MEMS Cap device, as there is no external alert that can function as a reminder.


\subsection{Helping Hand}
\subsubsection{Feature and System Architecture}
\begin{figure}[h]
    \centering
    \includegraphics[scale=0.5]{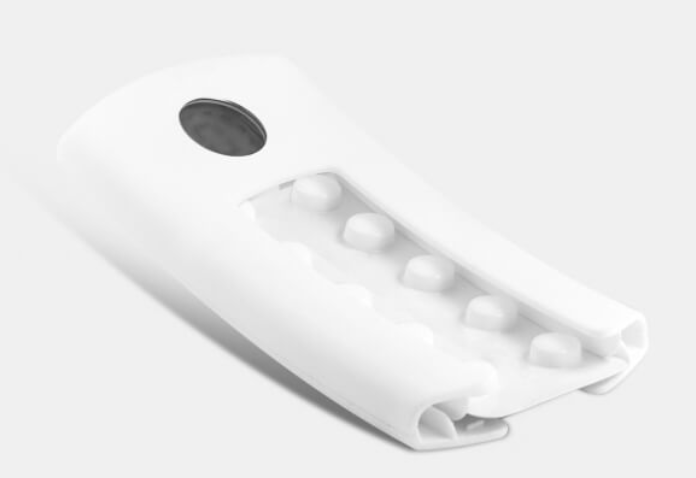} 
    \caption{Helping Hand (the blister sleeve) will detect the date and time when the blister is removed from the sleeve and reinserted, the blister pack must be in a particular dimension, the circular LCD will display the dosage within the last 24 hours and the time since the last dose [9].}
    \label{fig:HH}
\end{figure}

Helping Hand from B\&O Medicom shown in Fig. \ref{fig:HH} is a smart medication adherence monitor that uses a reusable blister sleeve to record the event of date and time when patients take the blister out of the sleeve [9]. Same as the MEMS Cap above, it can store up to 4,000 dosing events and the data stored can be read through MEMS reader and analyzed with MEMS ADHERENCE SOFTWARE. The data that was stored will transfer from the MEMS Cap to the Reader by wirelessly contactless communication[9]. The battery can support everyday use for up to 3 years and unlike MEMS Cap, Helping Hand comes with an LCD display that shows the number of the dose taken in the last 24 hours and the time passed since the last medication [9]. Helping Hand is restricted by the blister dimension, therefore one Helping Hand may only support one medicine, changing medicine will result in changing another Helping Hand where 3D printing was used to fit with the new blister [9].\\
\indent The overall system of Helping Hand can be separated into three parts, the Helping Hand blister sleeve, the blister pack and the MEMS Reader. Helping Hand is the blister sleeve that covers the blister pack, the integrated microcircuits inside the Helping Hand will detect for the event as if the blister is removed and reinserted in the sleeve. The stored events will be stored in the micro-controller inside Helping Hand and wait for the  MEMS Reader to extract the data. The MEMS Reader equip with the NFC reader and writer to communicate with the Helping Hand to read the stored event data [9].
\subsubsection{Performance Test}
The study of Helping Hand in [14] was based on user performance, satisfaction and acceptability, where 11 kidney transplant patients and 10 healthy volunteers were studied for 3 weeks. The performance was delivered through quantitative questionnaires and the satisfaction and acceptability were delivered through qualitative interviews and quantitative survey questions. The performance was evaluated in three areas: user error during operation, timing of operating tasks and thoughts on performing tasks. It was found that 57\% of the participants use the Helping Hand without error, most of the errors occur during the blister removal and insertion such when the blister was removed too fast that the event was not recorded and suggestions were made that the remove and reinserting mechanism is not friendly to elders. The satisfaction part involves many dimensions such as physical, privacy, human interaction, self-concept of dependency, routine and sustainability. It was found that 50\% of the participants think the device shape is not great, 66\% concerned about the privacy issue, 18\% of the patients resist sharing the collected information with physicians. More than 80\% of the participants think the device is providing a positive influence on adherence and less than 30\% will feel dependency on the device. Because of the inconvenience design and privacy issue, only 40\% of the participants would like to use Helping Hand in the future. Therefore, future technical improvement is required for Helping Hand but the main issue of one can only use one medicine in each Helping Hand will still remain.
\subsection{Smart Blister}

\subsubsection{Feature and System Architecture}
\begin{figure}[h]
    \centering
    \includegraphics[scale=0.5]{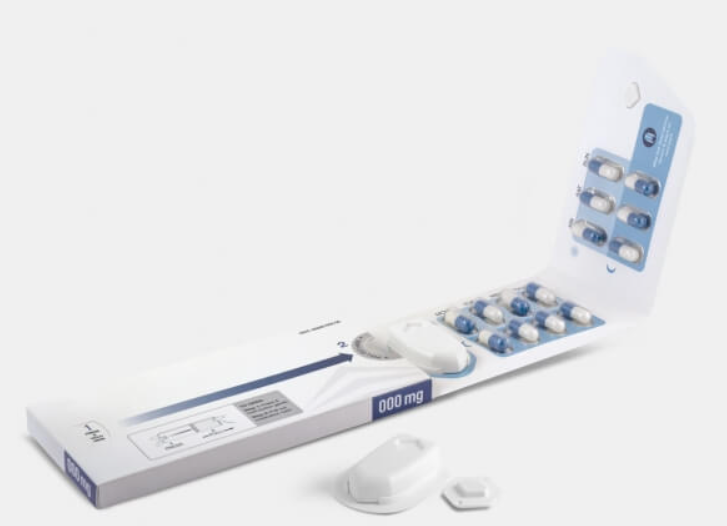} 
    \caption{Smart Blister equip with self-adhesive label (the block at the left of the blister) will detect the date and time when a pill is removed from the blister, the blister pack can be any blister pack available [9].}
    \label{fig:SB}
\end{figure}

Smart Blister from Qolpac BV, Eindhoven, different from the MEMS Cap and Helping Hand, is uses a RFID based self-adhesive label which contains event detection microcircuit that affixed on the existing blister pack [3]. According to the market existing product that can seen in Fig. \ref{fig:SB} [9], it records the date and time when the pill is taken out from the blister and the data stored can be read through MEMS reader and analyzed with MEMS ADHERENCE SOFTWARE. The data that was stored will transfer from the Smart Blister to the Reader by wirelessly contactless communication[9]. Ideally, it can be installed on any blister pack and is able to reuse on other blisters when the current one is empty. \\
\indent The overall system of Smart Blister was built based on the current available blister pack, the event detection microcircuit in the self-adhesive label can detect for the event when a pill is removed from the blister pack and such data (obtained from the detector inputs, a counter and a clock generator) can be acquired by NFC devices through NFC or RFID communication [3]. The system has no warning mechanism but it can record the dosage up to each count.
\subsubsection{Performance Test}
The test of Smart Blister was performed in [3] involved 20 pharmacies, where the patients will be offered the Smart Blister and monitor the adherence for 3 months. Each time the patient refill the prescription (14 pills per blister) at the pharmacy, the data recorded on the label will be transmitted to the system where data can be stored and managed. A total of 104 participants involved in this study of adherence and it was found that the overall adherence rate was 85\% and it will decrease with the period of followup with the pharmacy. 17\% of the evaluated use was found that recorded error events which may caused by breakage or nearby conductive tracks on the label [3]. As a conclusion, although Smart Blister is easy to use and elder friendly, the robustness of the label and the event detection need to be reinforced, there is also lack warning or any other visual/audible resource.

\subsection{Comparisons}
The comparison between the three products above is summarized as TABLE \ref{Tab:1}. It can be found that although each device has their own advantage such as MEMS Cap has the best robustness and accuracy on event detection, Helping Hand comes with an LCD display to provide dosage information, Smart Blister has the best privacy protection, but they all share or owns potential problems that will reduce the rate of adherence. What shall be achieved on the proposed design should take advantage of the existing products on: accurate event detection, warning or alarm display, long battery lifetime, easy to use data transmitting technique. While on the other hand improve on the areas that are considered to be a shortage in the existing products: real-time user feedback, ability to adapt with different medicines without changing the design, ability to detect the number of pills for each dosage.

\begin{table*}[ht]
\caption{Comparison between the MEMS Cap, Helping Hand and Smart Blister}
\begin{tabular}[width=\linewidth]{|l|l|l|l|} 
\hline
                                        & MEMS Cap     & Helping Hand       & Smart Blister                 \\ \hline
Event Detection Robustness and accuracy & High         & High               & Medium                        \\ \hline
Reminder/Alert                          & Optional LCD & LCD                & N/A                           \\ \hline
Battery Lifetime                        & 2-3 years    & 2-3 years          & N/A                           \\ \hline
Time Needs for Event Registration   & 3s                               & Blister out and back in          & Instant               \\ \hline
Privacy Protection                      & Medium       & Medium             & High                          \\ \hline
User Feedback                       & Require MEMS Reader and software & Require MEMS Reader and software & N/A                   \\ \hline
Event Storage                           & 4,000        & 4,000              & Depends on the blister (6-18) \\ \hline
Compatibility with Different Medicines  & Perfect      & Bad                & Perfect                       \\ \hline
Ability to detect for dosage number & N/A                              & N/A                              & Through data analysis \\ \hline
Support for Multiple Medicines          & N/A          & N/A                & N/A                           \\ \hline
Reusability                             & Reload pills & Change new blister & N/A                           \\ \hline
\end{tabular}
\label{Tab:1}
\end{table*}

\section{Design and Implementation}
Take the criteria that were developed from the comparison section and the initial idea, the objective of the device can be delivered into two portions in the device architecture, the hardware part and the software application part. In the hardware part, the device should equip with functions as follow: Microcircuit and sensor that measure the dosage accurately, communication device to transmit the data, battery that can power the device for 2-3 years based on everyday usage. The software part of the system should equip with functions as follow: communication device to receive the data, proper data processing and managing, ability to provide warning or alarm, real-time feedback based on received data. Other than these fundamental functions, by cooperating with hardware and software, the system can be able to detect the number of pills based on different kinds of medicines. Furthermore, the software can take the idea of IoT to upload the data to the cloud for data sharing with doctors and family members. \\
\indent The overall architecture of the designed system is shown in Fig. \ref{fig:System}. When the pill was taken out of the case, the load cell inside the Smart Pill Case will weigh the total weight of pills left inside the case, then the weight will be processed by the micro-controller and passed to an NFC tag by RFID writer, at this step, the duty of the Smart Pill Case part is accomplished. Now the user will open the software application on the SmartPhone, the application will start to detect the NFC tag and read the weight information that was written in the tag. The weight information will then be processed into dosage based on the recorded pill information and previous weight stored in the memory and provide a warning regarding the dosage. The weight will be stored in memory to calculate the next dosage and can also be sent to the cloud by the Internet for data sharing or managing. Here, the system will be separated into the hardware (Smart Pill Case) part and software (Smart Phone application) part and discuss the design and implementation in detail.

\subsection{Hardware System}
\subsubsection{Load Cell}
\begin{figure}[h!]
    \centering
    \includegraphics[width=\linewidth]{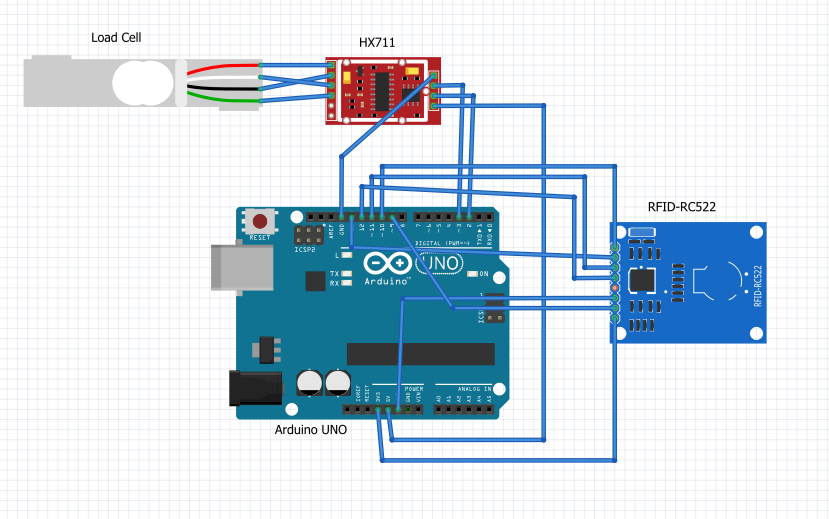} 
    \caption{The wiring diagram of the Smart Pill Case, the load cell is a 1KG portable load cell, the HX711 is the load cell amplifier, the RFID-RC522 is a 13.56MHz RFID module, they are all connected and controlled by Arduino Uno. }
    \label{fig:Wire}
\end{figure}

Three types of weight sensors were tested duing the sensor selection: 0.5 Inch diameter resistive thin-film pressure sensor, 10KG load cell and 1KG load cell. During testing, it was found that the resistive thin-film pressure sensor requires a relatively high starting force before it can read on weight changes, it might be a problem when only a few pills are left in the case. The other problem is that the output carried with noises as the resistor may be affected by temperature change, therefore it is not the best choice in this project. Compare the 10KG load cell with the 1KG load cell, the 1KG load cell is able to provide more accurate weight for lightweight objects, therefore the load cell that I chose is the AREEY470 1KG Portable Electronic Scale with the HX711 load cell amplifier. The wiring is shown in Fig. \ref{fig:Wire}. The two components belong to the physical layer of the system, the amplified weight information inputs to the micro-controller by a non-I$^2$C compliant two-wire protocol, the two wires ( pin2, pin3 in Fig. \ref{fig:Wire}) are output pin and clock pin that is used to transmit the weight data and check if the micro-controller is ready to process incoming data. Using the HX711 library and calibrate the calibration factor with known weights, the weight of an object can be acquired.

\begin{figure}[h!]
    \centering
    \includegraphics[scale=0.14]{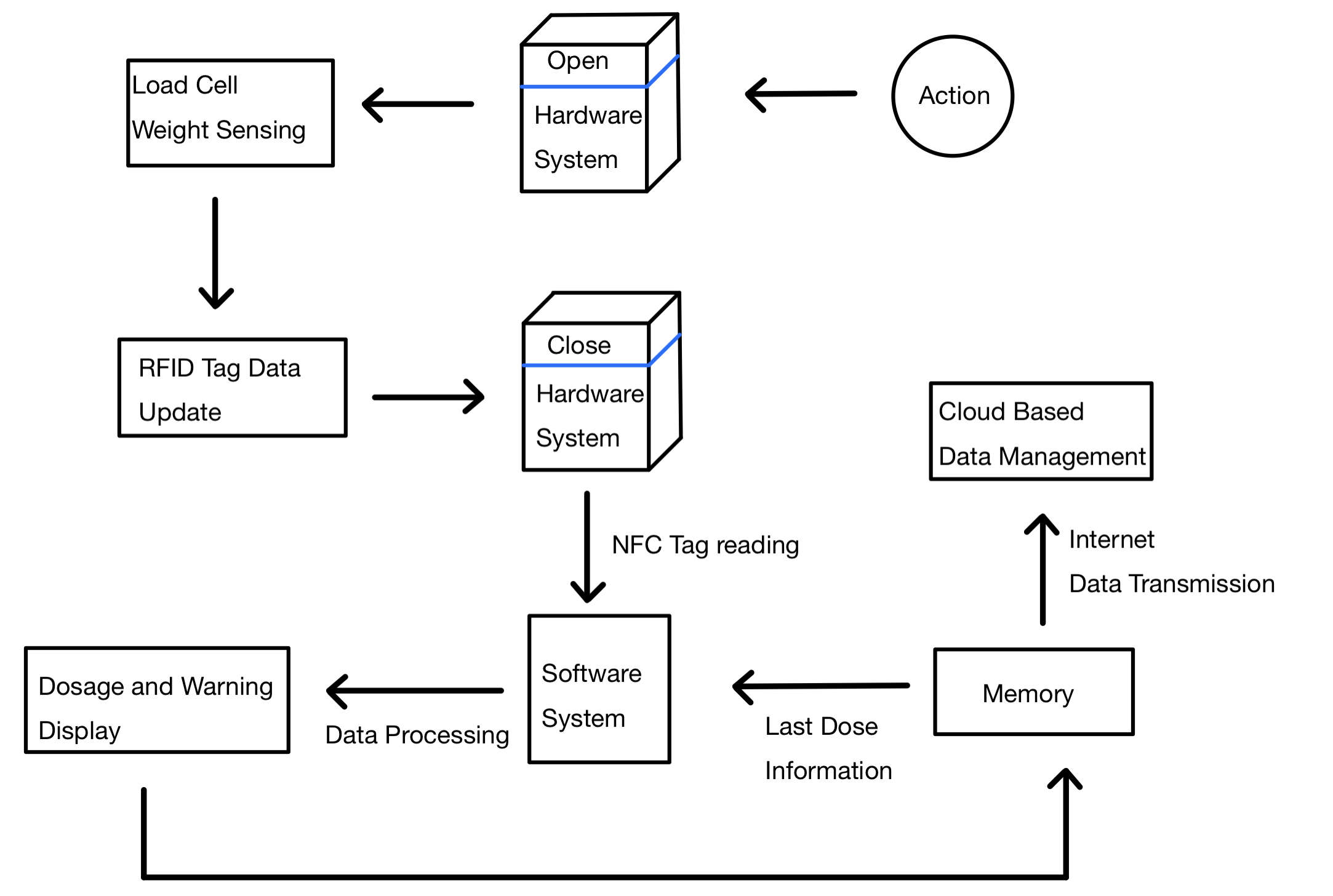} 
    \caption{The overall information flow of the system, the action refers to the action of taking pills out of the case, the upper hardware part occurs in the Smart Pill Case and the lower software part occurs in the Smart Phone.}
    \label{fig:System}
\end{figure}
\subsubsection{RFID}
Radio-frequency identification (RFID) is a wirelessly contactless communication technology that uses radio frequency waves to transmit data. Between an RFID device and the micro-controller, the communication was under Serial Peripheral Interface (SPI) protocol which has a clock pin (CLK) to synchronous the communication; a Master Out Slave In pin (MOSI) that transmits the data from the Master side (the side that generates the clock, often the micro-controller) to the Slave side (often the sensor) to request for data; a Master In Slave Out (MISO) pin that transmits the required data from the sensor to the micro-controller; a slave select (SS) pin that selects the sensors that share the same bus for which the micro-controller want to communicate with [15]. Between the NFC tag (under ISO 14443-3A standard) and the NFC reader in Smart Phone, the information was transmitted using a 13.56 MHz electromagnetic field under the protocol defined in ISO/IEC 14443-3:2011 Part 3 Type A. The data is in the NFC Data Exchange Format (NDEF).  \\
\indent In this project, RFID was chosen as the communication technique between the Smart Pill Case and the Smart Phone for several reasons. Firstly, characterized by the smart health adherence application, the data that needs to be analyzed are meta data, therefore, the size of the data that need to be transmitted is small. For the Smart Pill Case, only the weight of the pills left need to be saved and transmitted, while RFID is a great choice for small size data communication. Secondly, RFID provides seamless communication between different devices, compare to Bluetooth, RFID (NFC) needs only a single touch to complete the transmission while Bluetooth takes time to pair between devices. Thirdly, the Smart Pill Case was designed to save the information inside the NFC tag memory, therefore, the system can be powered only during the time when the lid of the Smart Pill Case is on. During this period, the load cell and RFID will keep updating the weight value in the NFC tag memory and when the lid is off, the power is cut, left only the NFC tag with the latest weight information to be accessed by the Smart Phone application. In such a way, the battery life of the device can be optimized.\\
\indent MFRC522 was used as the RFID device that writes the weight in the NFC tag memory, the wiring diagram is shown in Fig. \ref{fig:Wire}. By modify the MFRC522 library "ReadandWrite" example, the data can be written into 16 bytes array and stored in the memory blocks that are available for data storage. Following the plain text format regulated under NDEF, some bytes in the first data block (block 4) were fixed so that the NFC reader will output the data into plain text format. As the pill weight never exceeds 100g, taking one decimal place, four bytes in total will be used in the micro-controller to represent the weight in the NFC tag. Three variable bytes will represent the weight and one fixed byte will represent the decimal dot. Each time the weight is updated, the three variables that represent the weight will be changed into the corresponding weight number in ASCII and then convert to hex to fit into the memory space of the NFC tag. When the NFC reader scans the NFC tag, the text in the format of "xx.x" will pop up which shows the current weight in grams of the pills left in the case.
\subsubsection{Arduino Uno}
The Arduino Uno is the micro-controller that was used in this project. It was used to process the incoming weight data from the load cell and the amplifier to address the current weight of the pills left in the case, then the weight will be coded into hexadecimal as the RFID section discussed above, the encoded information will be passed through the RFID writer and record in the memory of the NFC tag. The Arduino Uno is powered by a 9V battery connect with the lid switch so that when the lid is open, the system will start to scan the weight and record it on the NFC tag. When the pills were taken out from the case and the lid was off, the last data recorded on the tag will be the current weight of the pills left and the system will turn off until the next dosage for battery saving. The source code for the Arduino Uno that runs the Smart Pill Case can be found in {\it https://github.com/MibclAric/Smart-Pill-Case/blob/main/Project.ino}
\subsubsection{Smart Pill Case Prototype}
\begin{figure}[h]
\centering
    \includegraphics[width=\linewidth]{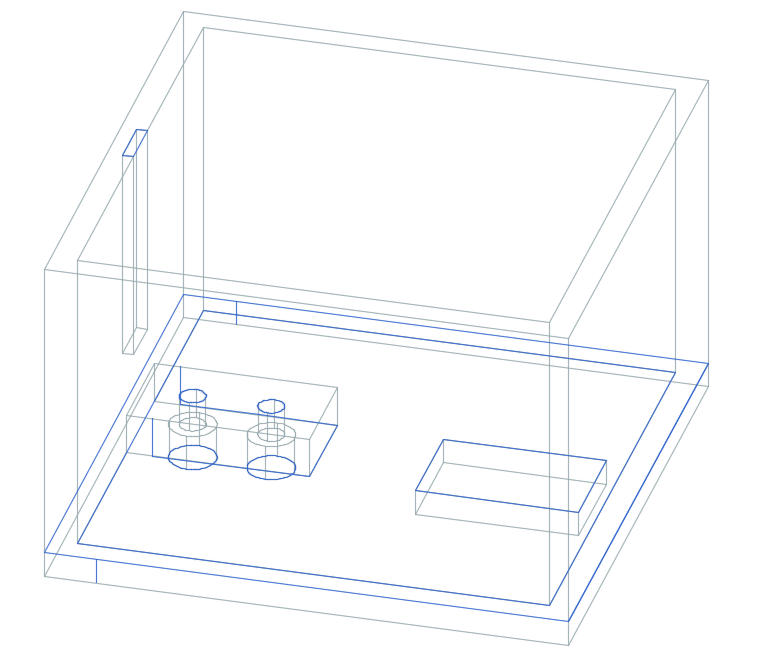} 
    \caption{The base part of the Smart Pill Case, the left block with holes inside the base carries one side of the load cell, the pill container is mount on the other side of the load cell. The notch at the left is for wiring.}
    \label{fig:Base}
\end{figure}

The mechanical parts of the Smart Pill Case was designed in NX. The case was designed to provide space for the pill container, the micro-controller, the load cell and the RFID chip. \\
\indent The base of the Smart Pill Case shown in Fig. \ref{fig:Base} provides room for the load cell and the pill container. One side of the load cell will be mount on the block stage with screws and the pill container will be mount on the other side. This design was made to ensure there is enough room left for the load cell to bend as the weight (force) is measured by measuring the change in resistance which was resulted from the bending of the load cell. The load cell wires will go through the notch towards the micro-controller at the lid.\\

\begin{figure}[h]
    \centering
    \includegraphics[width=\linewidth]{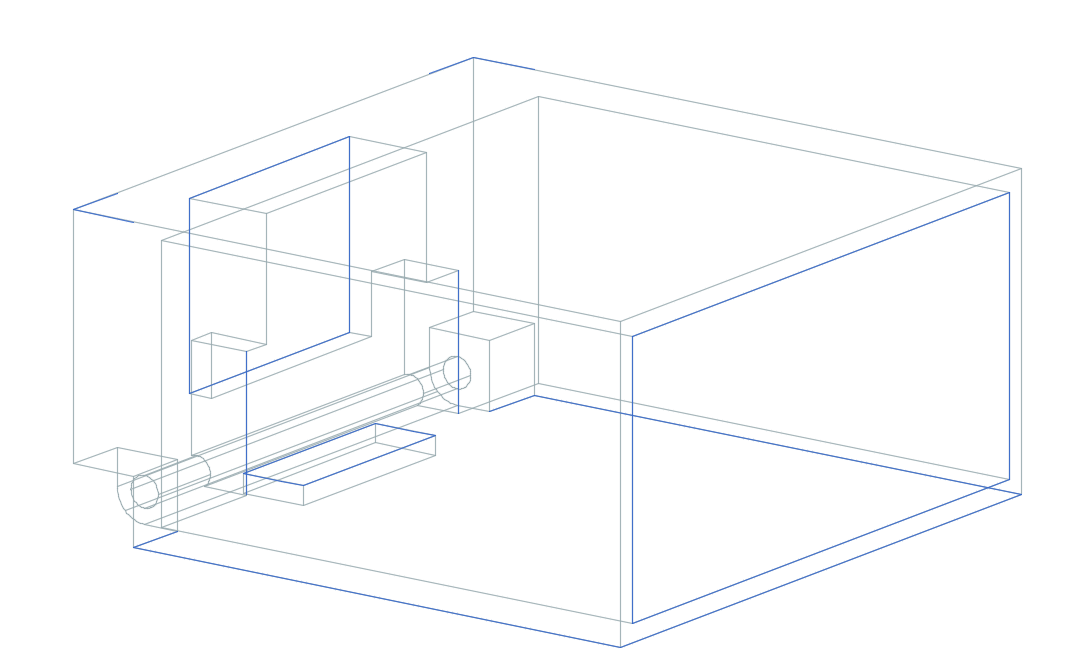} 
    \caption{The lid part of the Smart Pill Case, the micro-controller, battery and MFRC522 will be fit inside the lid.}
    \label{fig:Lid}
\end{figure}

\indent The lid of the Smart Pill Case shown in Fig. \ref{fig:Lid} provides room for the micro-controller, MFRC522 and battery. The micro-controller is at the bottom inside the lid and MFRC522 sits on the micro-controller. The NFC tag is attached at the top of the MFRC522 and in this way it can be read directly by tap the top of the lid with the Smart Phone.\\

\begin{figure}[h]
    \centering
    \includegraphics[width=\linewidth]{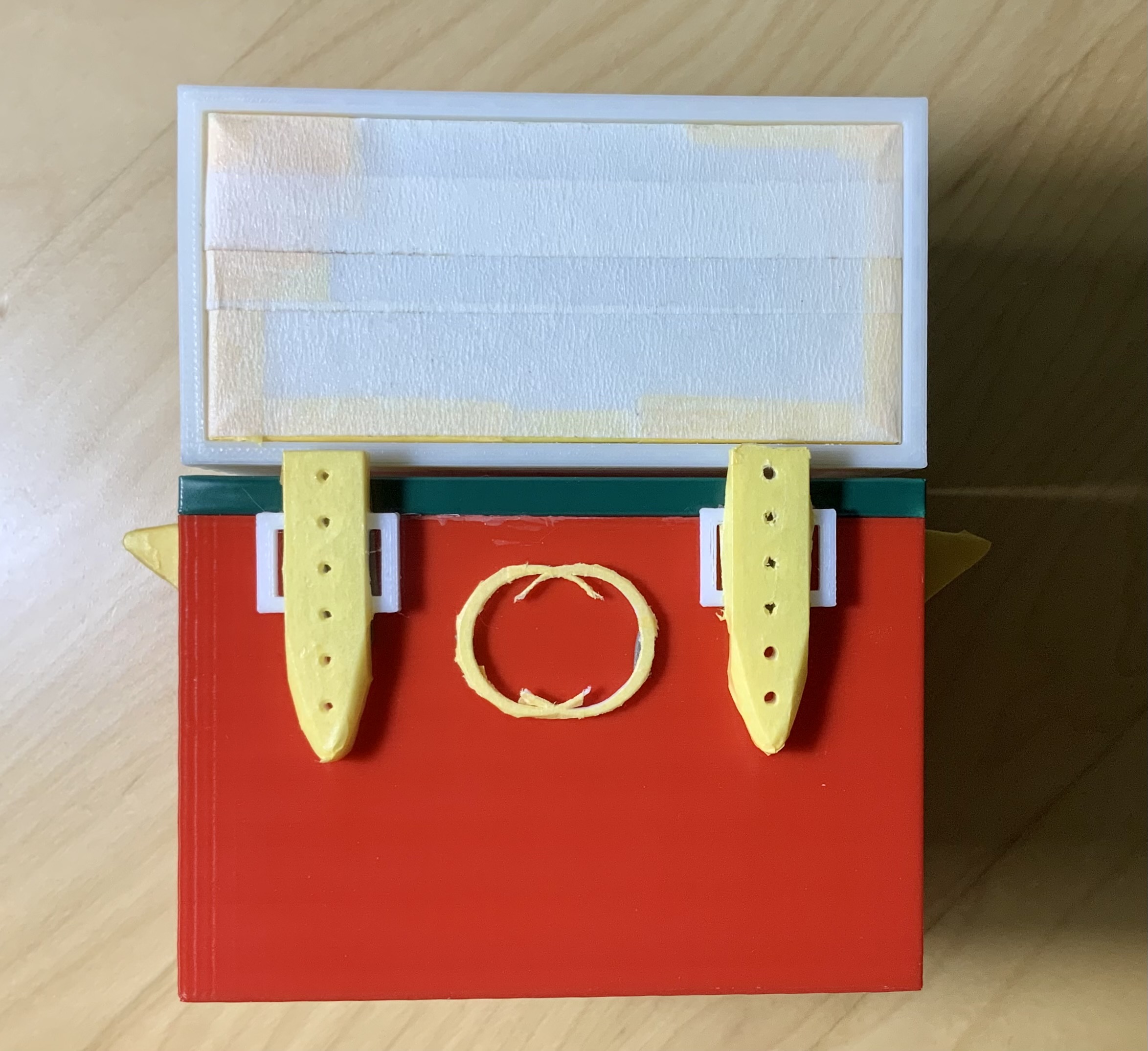} 
    \caption{Front view of the Smart Pill Case, the white part is the lid and the red part is the base.}
    \label{fig:Front}
\end{figure}

\begin{figure}[h]
    \centering
    \includegraphics[width=\linewidth]{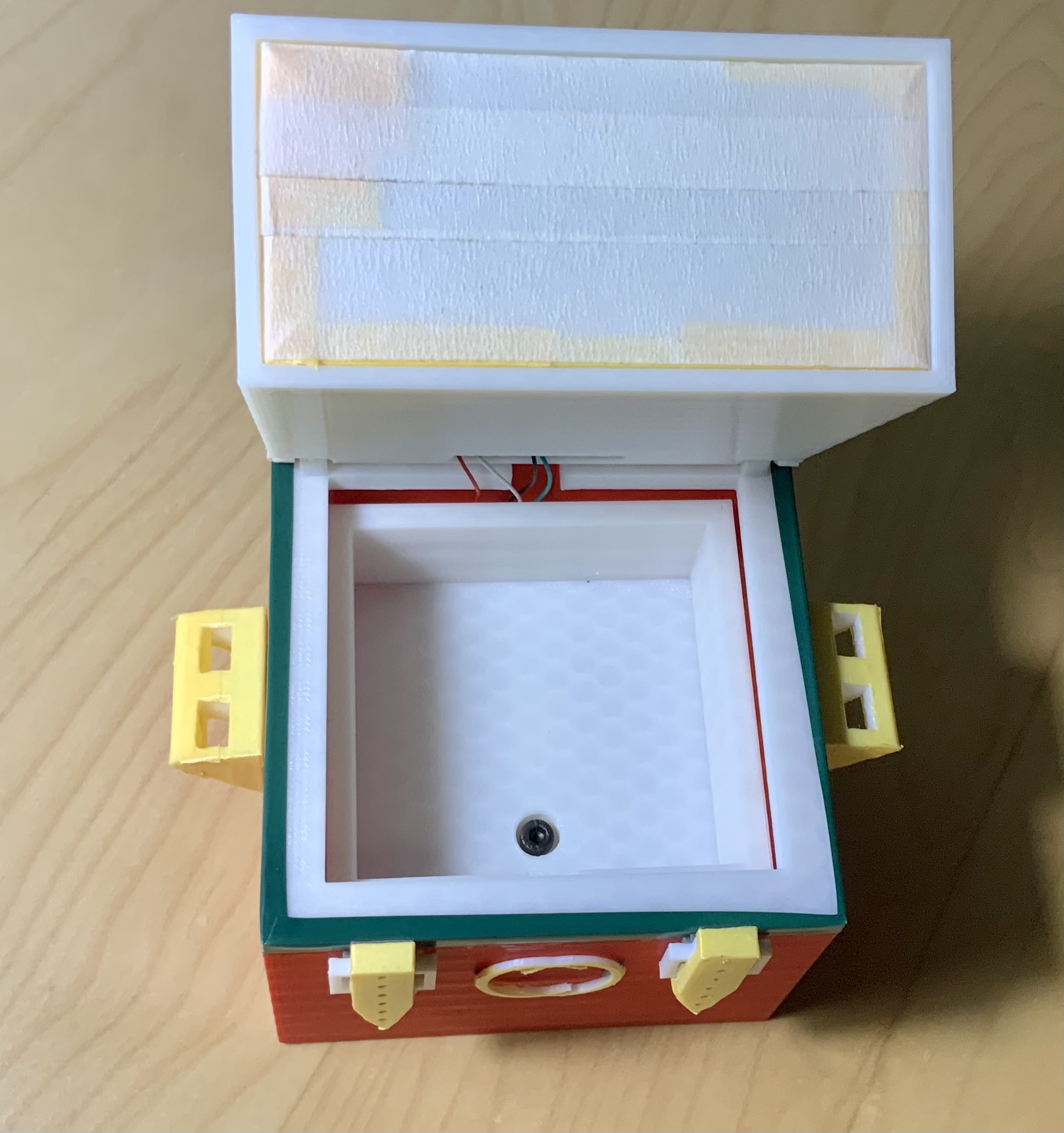} 
    \caption{Inner view of the Smart Pill Case, the white block inside the base is the pill container that store the pills.}
    \label{fig:Inside}
\end{figure}

\begin{figure}[h]
    \centering
    \includegraphics[width=\linewidth]{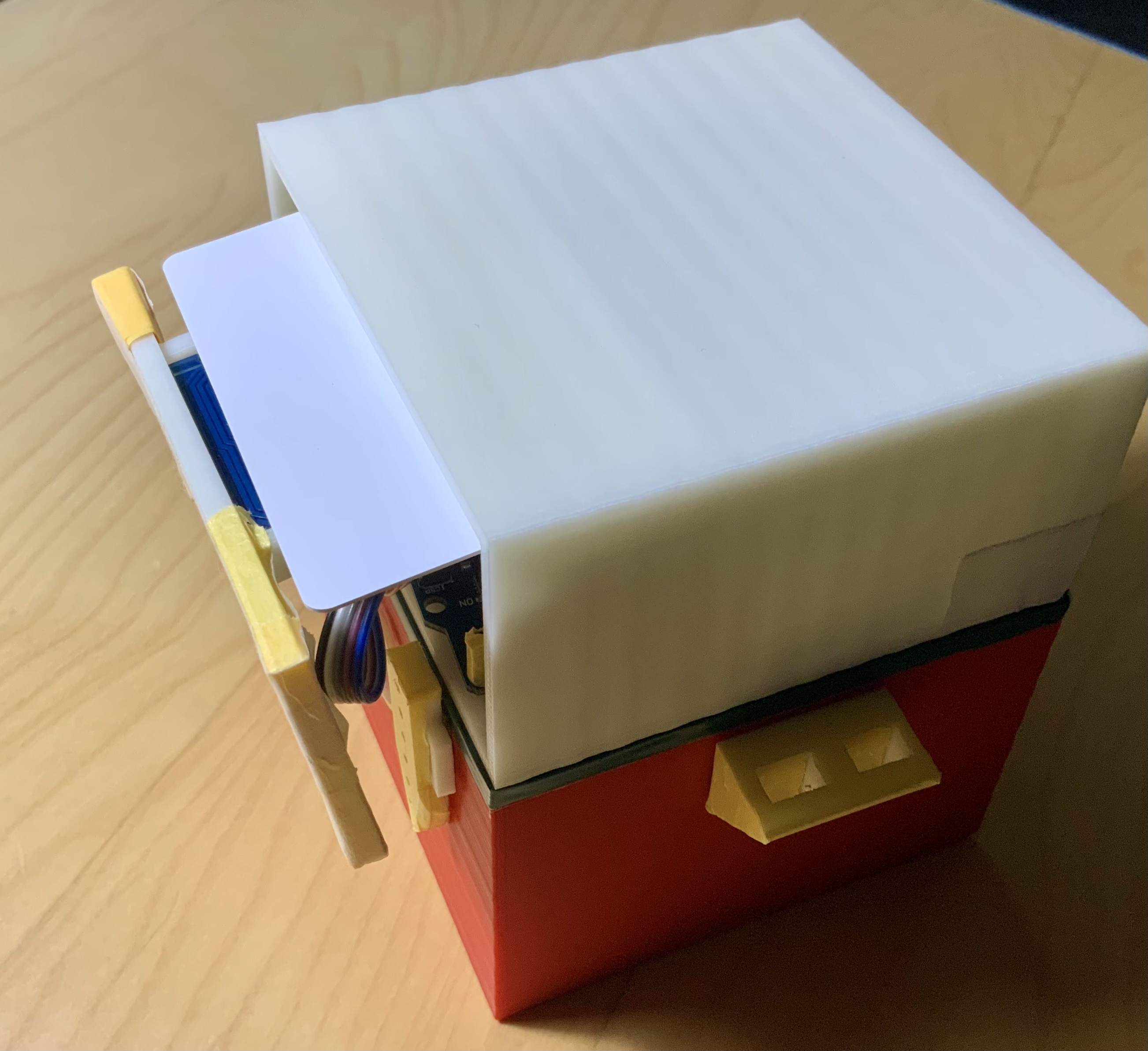} 
    \caption{The micro-controller, MFRC522 and battery are inside the lid. The lid can be closed with a front cover. The white card is the NFC tag that record the data, the data can be read by tap the top of the lid with a Phone}
    \label{fig:LidI}
\end{figure}

\indent The prototype was 3D printed and assembled as Fig. \ref{fig:Front}, Fig. \ref{fig:Inside} and Fig. \ref{fig:LidI} shown. In summary, the Smart Pill Case is the physical device of this health adherence tool, the user will first open the lid and take medicine out from the pill container, during this procedure, the load cell and the MFRC522 will continuously update the current pill weight and store the data in the memory block of the NFC tag \cite{hamidi2019systems}. When the user successfully took the medicine out and closes the lid, the power will be cut and the NFC tag now is stored with the weight of the pills left in the pill container after some pills were taken out. A Smart Phone can then tap on the top of the lid to receive the weight data for further processing. This part will be discussed in the next section.

\subsection{Software System}
\begin{figure}[h]
    \centering
    \includegraphics[width=\linewidth]{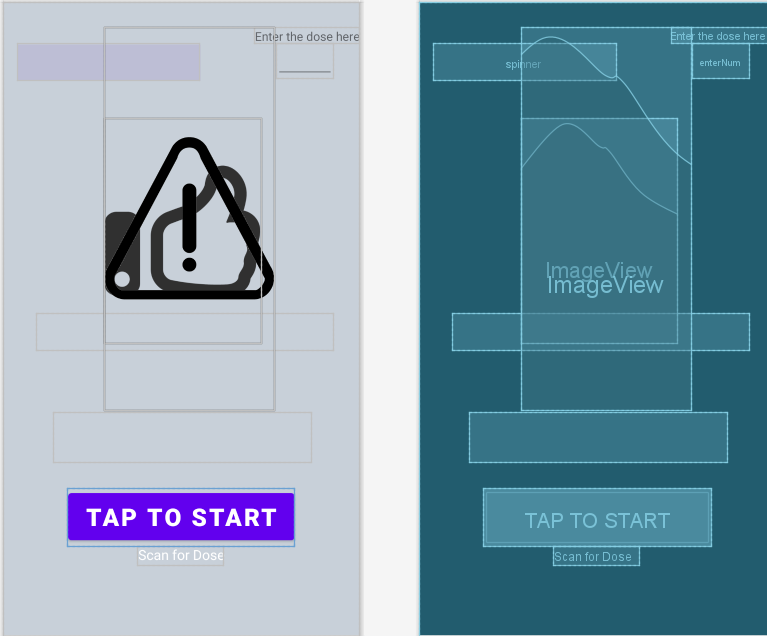} 
    \caption{The User Interface of the Smart Pill Case App. }
    \label{fig:UI}
\end{figure}

\begin{figure}[h]
    \centering
    \includegraphics[scale=0.1]{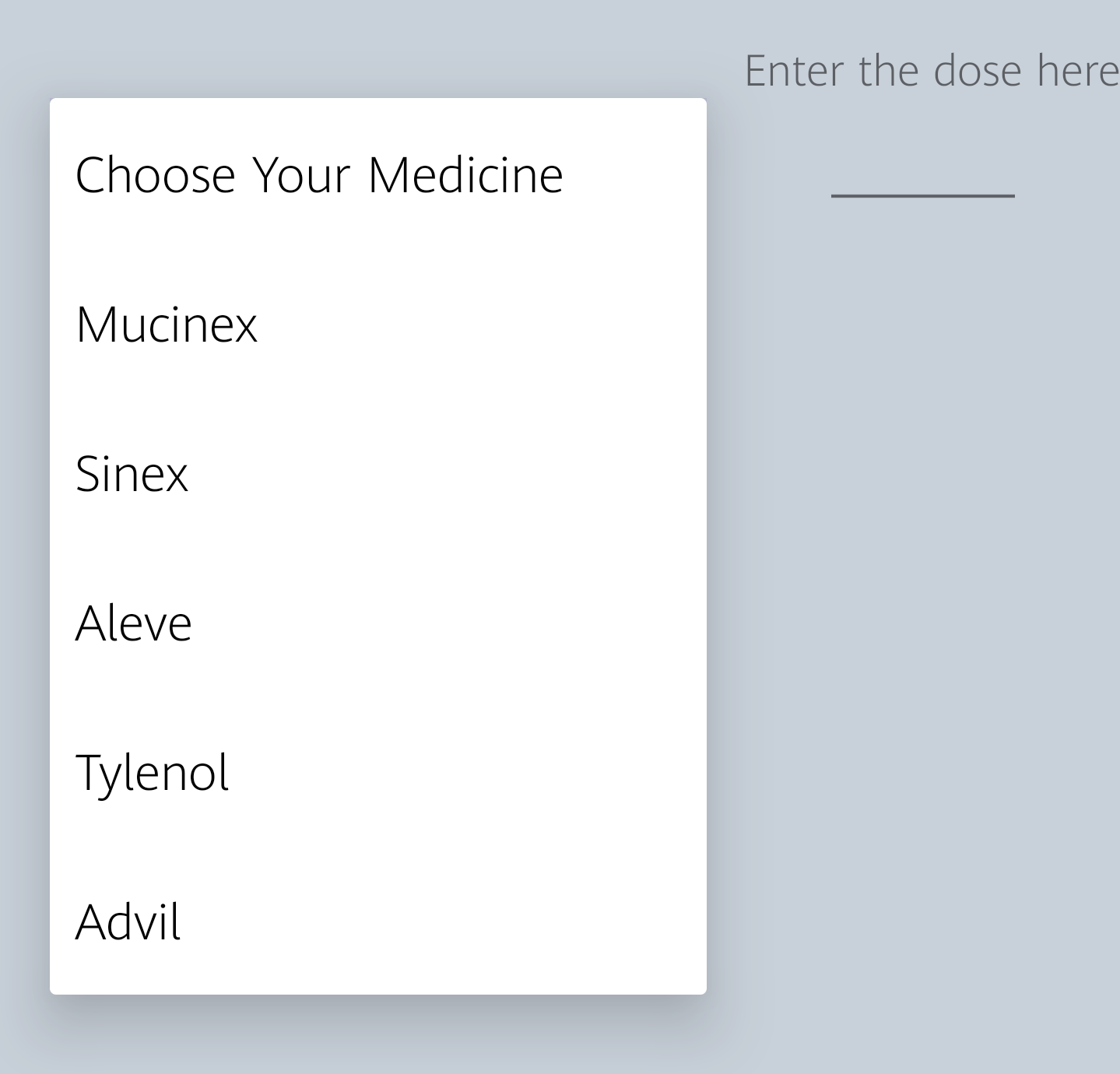} 
    \caption{The top area of the APP will provide input setting from user based on prescription. }
    \label{fig:Top}
\end{figure}

As Fig. \ref{fig:System} shows, the software system will read the weight from the NFC tag for analysis, these tasks can be accomplished by software applications. Here, Android Studio was used to design such an application that provides a user interface and a background algorithm to take prescription information from the user, scan the NFC tag to acquire the data and then process the data to provide real-time feedback to the user.\\
\indent The UI architecture of the APP is shown in Fig. \ref{fig:UI}, it can be separated into three areas by the functionality it provides. The top area is constructed by two user inputs, one is a spinner at the left and the other one is an input text at the right. The spinner provides a list of selections, press the spinner will display a dropdown menu and the selected variable will be saved at the back, here, the type of medicine will be displayed as Fig. \ref{fig:Top} shows, once a medicine was selected, the scanned dose number will be calculated using the unit weight of this selected medicine (the unit weight of the medicines were predefined in the algorithm). The input text at the right will record the recommended dosage from the prescription by the user, it will be used to determine if the user is taking an extra amount, insufficient amount or the correct amount by comparing the recommended dosage with the current scanned dosage.\\

\indent The second area is the area in the middle of Fig. \ref{fig:UI}, it displays information to the users. There will be two text blocks and two image blocks that can be displayed in this area, one of the text blocks will be showing the current dose number and the other text block will be showing the context information such as warning. The two images are the thumb-up image and the warning image, when the scanned dose number equals the recommended dosage, the thumb-up image will be displayed and when the dose numbers are not equal, meaning the user is taking a wrong amount of medicine than what is needed, the warning image will be displayed.\\
\indent The third area is the area at the bottom of Fig. \ref{fig:UI}, it is constructed with a button and when the button is being pushed, the two user input variables will be updated. By saying so, if either one of the user inputs was changed, by tap the button, the information at the back stage will be updated and the next scanned dose number including the warning will be based on the new information.\\
\indent The most important part of the application is to acquire the weight information from the NFC tag memory and use it to calculate how many pills (the number of doses) the user has taken. In MainActivty.kt, after the "TAP TO START" button is pushed, the APP will block the usage of NFC from other applications and only performing the tag detection. Once an NFC tag is detected and is connected with the Smart Phone, the NDEF message will be extracted from the NFC tag memory as a ndefMsg, then taking only the memory block that stores data and leave only the weight data bytes left, by converting the bytes back into integers, the weight information can now be available. Use the previous weight that was stored in the system minus the current weight and divided by the unit weight of this medicine, the number of doses that was taken can be calculated and displayed on the screen. The pseudo code is shown in Algorithm \ref{alg:NFC}.

\begin{algorithm}
\caption{A pseudo code for the NFC tag weight data extraction and processing}\label{alg:NFC}
\begin{algorithmic}
\Require NfcTagFound = 1
\Ensure NfcTag = connected
\State MedicineUnitWeight = SelectedMedicineUnitWeight
\State NdefMsg = ExtractNdefMsg
\State Value1 = NdefMsg[Value1.index]
\State Value1 = Value1.ConvertToDecimal
\State Value2 = NdefMsg[Value2.index]
\State Value2 = Value2.ConvertToDecimal
\State Value3 = NdefMsg[Value3.index]
\State Value3 = Value3.ConvertToDecimal
\State Weight = Value1$\times 10$ +Value2 +Value3$\times 0.1$
\State DoseNumber = (Weight - PreviousWeight) / MedicineUnitWeight

\end{algorithmic}
\end{algorithm}

\section{Result and Discussion}
The Smart Pill Case along with the developed software was tested in two aspects: the correct reading of pill numbers that were taken out from the Smart Pill Case in the App and the correct feedback the APP can provide based on the dose number. The quarter coin was used to replace a pill in this test section (they will be treated as pills in the following discussion), firstly, the weight of the pill (quarter coin) will be measured through an experiment. A total of 9 pills were put inside the Smart Pill Case and one pill was taken out each time, the weights that read from the load cell were recorded in TABLE \ref{Tab:2}, the Pills column represent the number of pills left inside the case and UW represent the calculated unit weight of the pill. It can be found that the two trials of the testing showed similar results and the unit weight of the quarter coin is between 4.4g to 4.5g (the real weight is 4.4g), this number was then recorded into the algorithm that if Tylenol was chosen (the quarter coin here represent the Tylenol pill), the number of the pills taken by the user will be calculated based on this recorded unit weight.

\begin{table}[h]
\caption{Experiment to Measure the Unit Weight of the Pill}
\begin{tabular}[width=\linewidth]{|c|c|c|c|c|c|}
\hline
Pills & Test1 & Test2 & Test1 UW & Test2 UW & Average UW \\ \hline
9     & 39.6  & 40.2  & N/A      & N/A      & N/A        \\ \hline
8     & 35.2  & 35.7  & 4.4      & 4.5      & 4.45       \\ \hline
7     & 30.7  & 31.3  & 4.5      & 4.4      & 4.45       \\ \hline
6     & 26.2  & 26.9  & 4.5      & 4.4      & 4.45       \\ \hline
5     & 21.7  & 22.4  & 4.5      & 4.5      & 4.5        \\ \hline
4     & 17.2  & 17.9  & 4.5      & 4.5      & 4.5        \\ \hline
3     & 12.8  & 13.5  & 4.4      & 4.4      & 4.4        \\ \hline
2     & 8.3   & 9.0   & 4.5      & 4.5      & 4.5        \\ \hline
1     & 3.9   & 4.5   & 4.4      & 4.5      & 4.45       \\ \hline
0     & -0.5  & 0     & 4.4      & 4.5      & 4.45       \\ \hline
\end{tabular}
\label{Tab:2}
\end{table}

\begin{figure} [!h]
\centering
\begin{subfigure}[b]{.5\linewidth} 
  \centering
  \includegraphics[width=\linewidth]{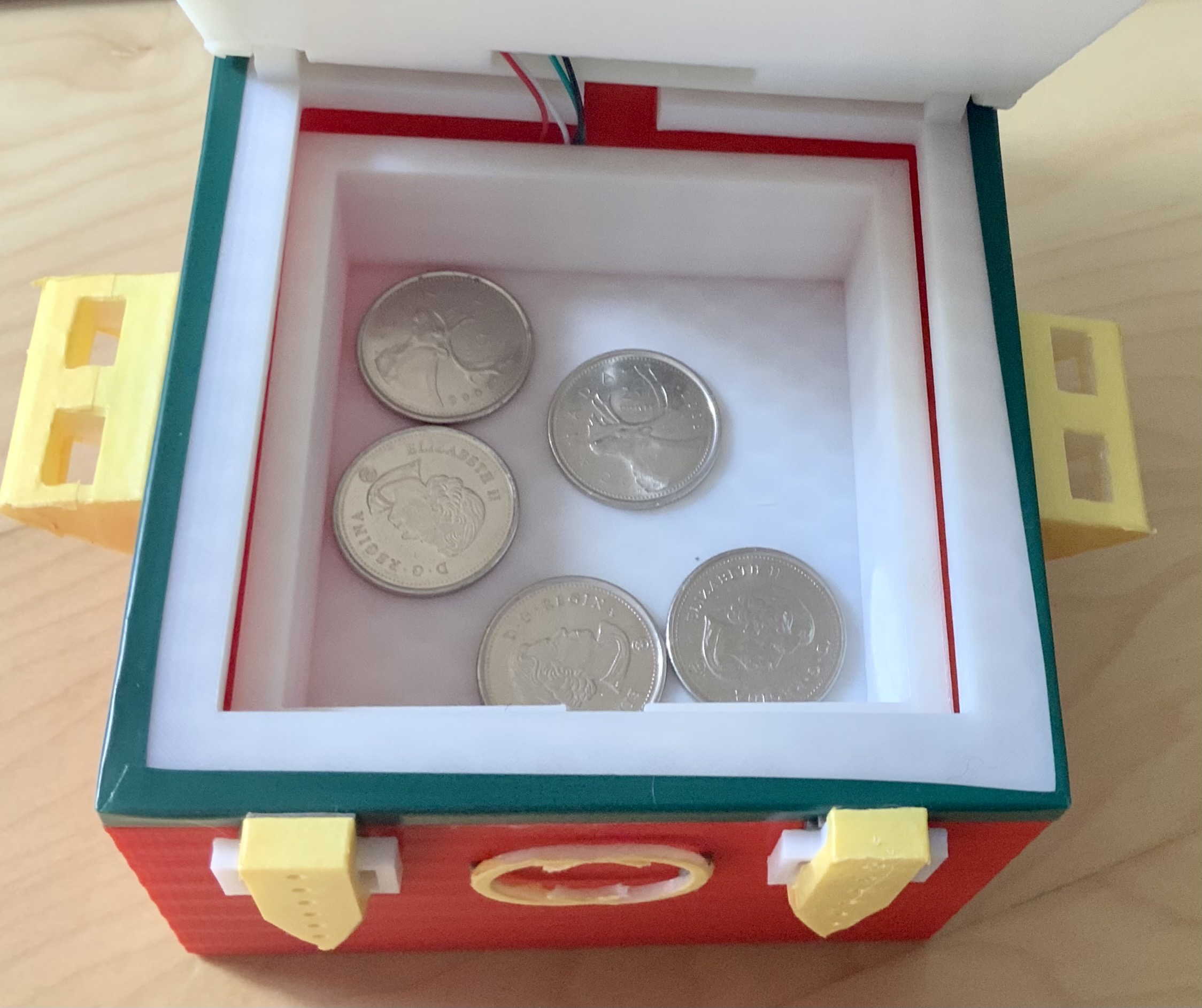}
  \caption{Pills (Coin) Left inside the Smart Pill Case}
  \label{fig:sub1}
\end{subfigure}%
\begin{subfigure}[b]{.5\linewidth}
  \centering
  \includegraphics[width=\linewidth]{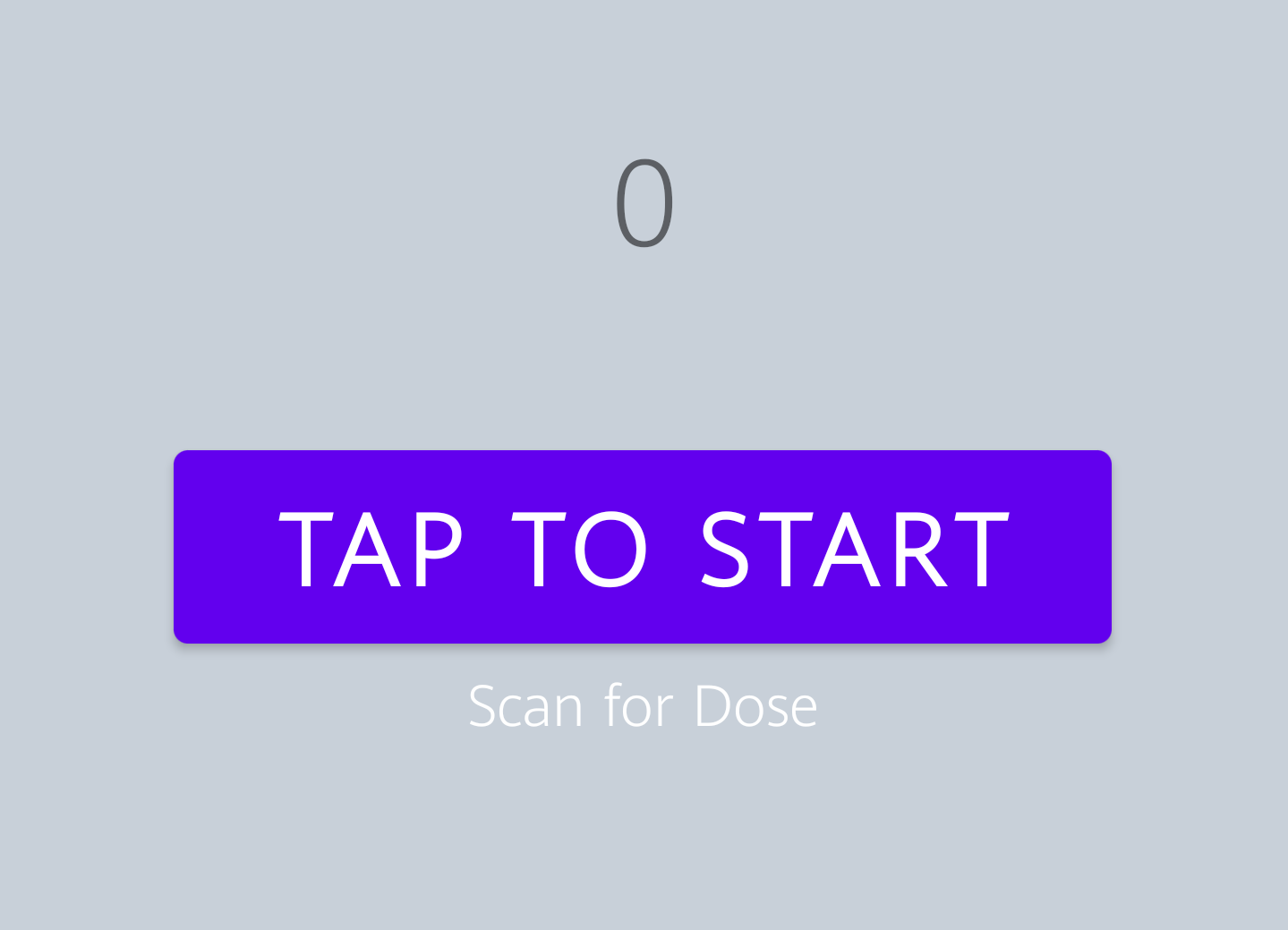}
  \caption{Pills Taken that the software read}
  \label{fig:sub2}
\end{subfigure}
\caption{Initial State of the Smart Pill Case and the APP with 5 coins inside.}
\label{fig:1}
\end{figure}

To test the system if the number of pills taken out can be correctly displayed on the screen, initially, 5 Tylenol pills were put inside the Smart Pill Case and the application was initialized by choosing the Tylenol as the medicine (the unit weight of quarter coin is recorded as Tylenol) and the initial weight was calibrated, the Smart Pill Case and the displayed dose number reading is shown in Fig. \ref{fig:1}.

\begin{figure} [!h]
\centering
\begin{subfigure}[b]{.5\linewidth} 
  \centering
  \includegraphics[width=\linewidth]{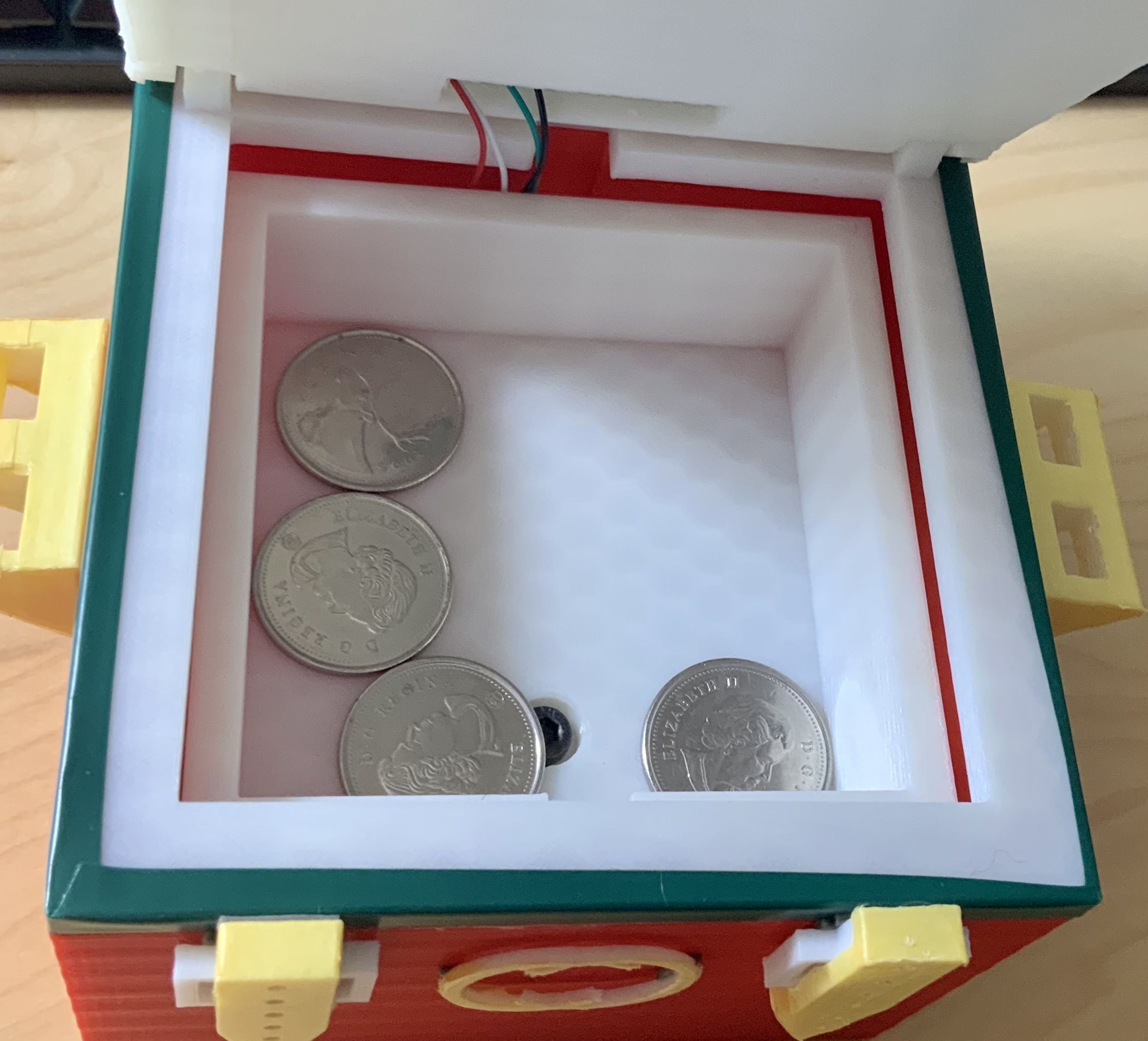}
  \caption{Pills Left inside the Smart Pill Case}
  \label{fig:sub1}
\end{subfigure}%
\begin{subfigure}[b]{.5\linewidth}
  \centering
  \includegraphics[width=\linewidth]{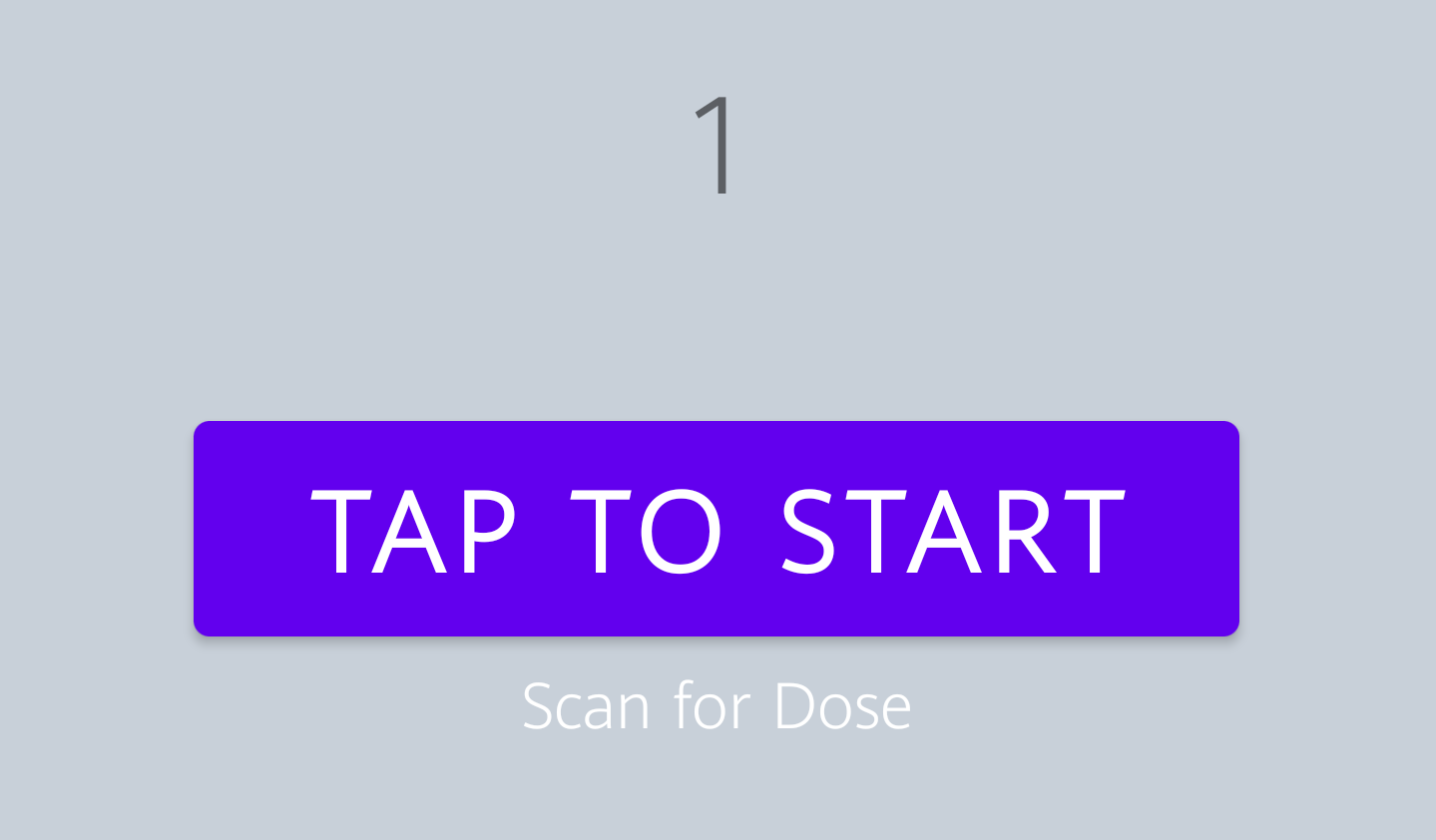}
  \caption{Pills Taken that the software read}
  \label{fig:sub2}
\end{subfigure}
\caption{One pill was removed from the Smart Pill Case and the Phone was tapped on the lid again.}
\label{fig:2}
\end{figure}

\indent Then the lid was opened and one pill was taken out from the Smart Pill Case. The next tap was made after the lid was closed and the result is shown in Fig. \ref{fig:2}. It can be seen that the system correctly calculated the number of pills that was taken out.

\begin{figure} [!h]
\centering
\begin{subfigure}[b]{.5\linewidth} 
  \centering
  \includegraphics[width=\linewidth]{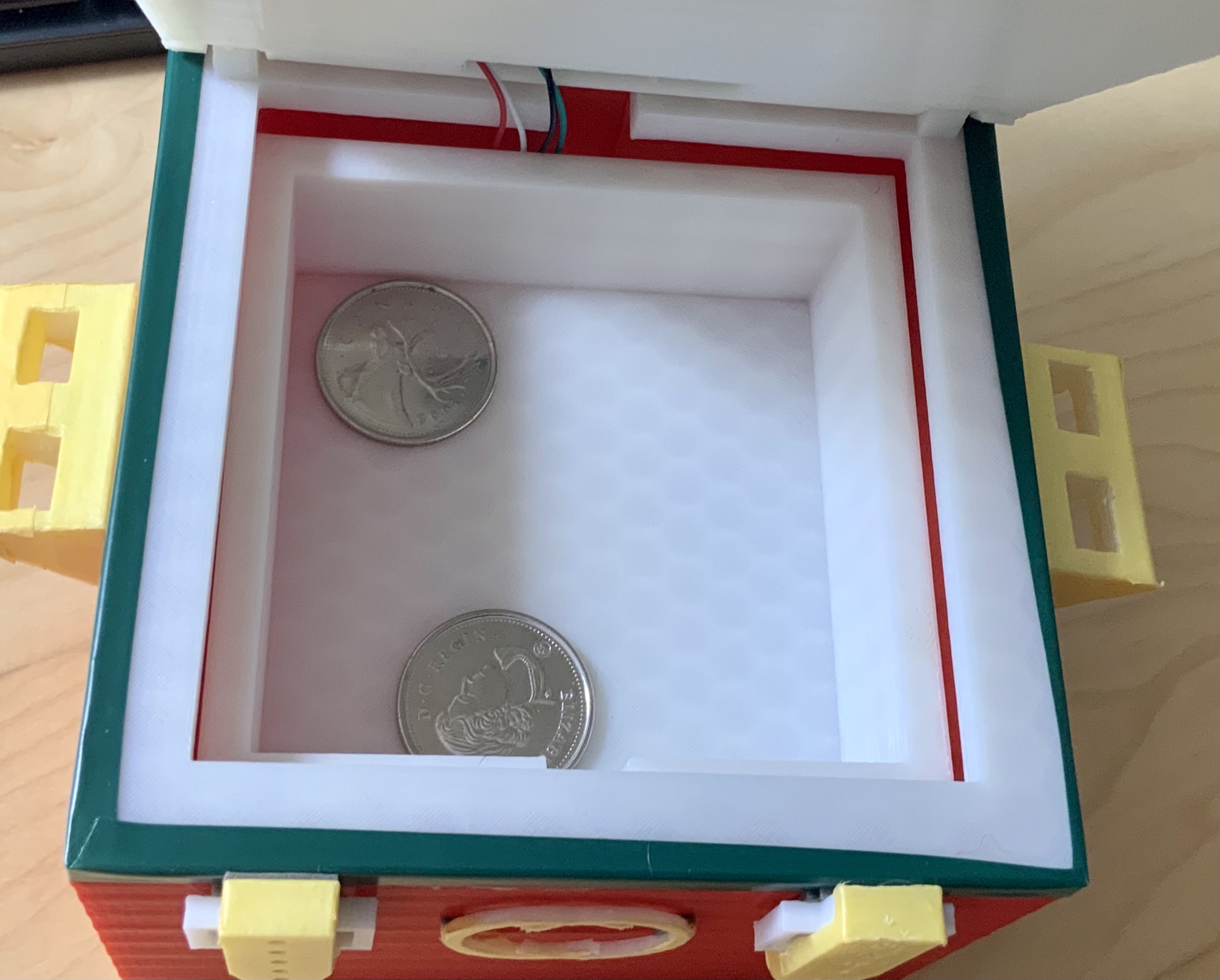}
  \caption{Pills Left inside the Smart Pill Case}
  \label{fig:sub1}
\end{subfigure}%
\begin{subfigure}[b]{.5\linewidth}
  \centering
  \includegraphics[width=\linewidth]{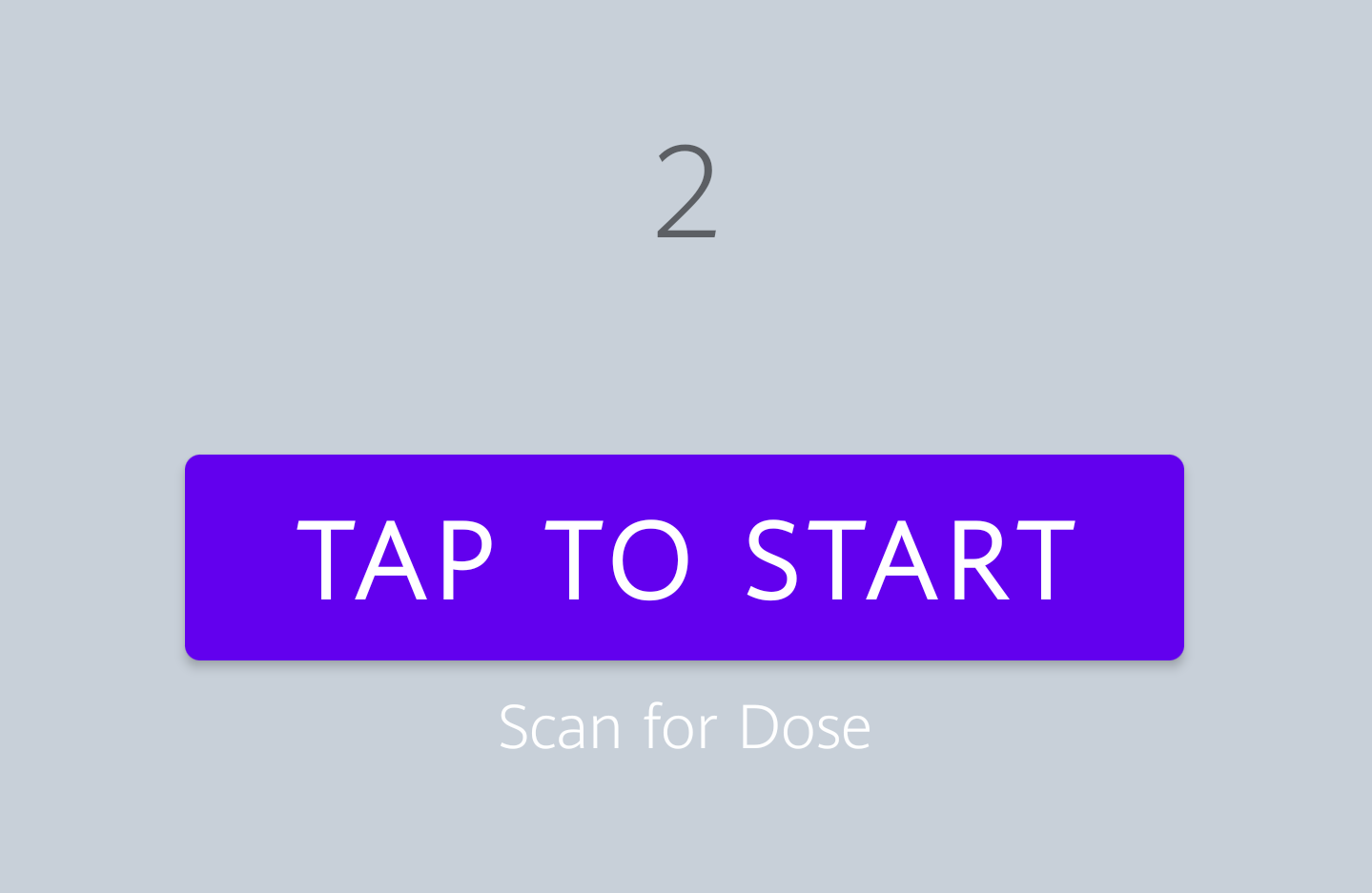}
  \caption{Pills Taken that the software read}
  \label{fig:sub2}
\end{subfigure}
\caption{Two pill was removed from the Smart Pill Case and the Phone was tapped on the lid again.}
\label{fig:3}
\end{figure}

\indent Then two pills were taken out with the same procedure and the result is shown in Fig. \ref{fig:3}. It can be seen that the system successfully calculated that two pills were taken compare with the last state rather than the initial state.

\indent As the test shown above, the Smart Pill Case with the APP can successfully detect how many pills were taken out. The next test will be focus on the function to provide warning. Ideally, if the detected dose equals to the preset recommend dose, then a thumb-up image will display, if they are not equal, a warning image will display and a text will be shown if the user is taking an exceed amount or an insufficient amount. Using the same initial state (5 pills), set the medicine to Tylenol (quarter coin) and the recommend dosage to 2, the screenshots in Fig. \ref{fig:4} shows how the system react if 1, 2 and 3 pills were taken out. It can be found that when 2 is the recommend dosage, taking 1 pill out will result the warning as Fig. \ref{fig:4} (a) shows and the text block warns that the user is taking 1 less than what should. Taking 2 pills away is the correct action, therefor the thumb-up image showed up in Fig. \ref{fig:4} (b). Taking 3 pills out will also result the warning as Fig. \ref{fig:4} (c) shows and the text block warns that the user is taking 1 more than what should.
\\At this point, all the fundamental function of the Smart Pill Case was proven to be working successfully. Next the battery life of this device will be discussed.

\indent The smart Pill Case uses one 9V 300mAh Lithium-ion Battery. The battery usage of each electrical component is shown in TABLE \ref{Tab:3}. Assume a user will open the Smart Pill Case 3 times a day and took 5 seconds each time to take the pills out, in this scenario, the battery can support the device for up to 5 years which is much more than what was required (2-3 years).

\begin{table}[h]
\caption{Power usage of each electrical components in Smart Pill Case}
\begin{tabular}[width=\linewidth]{|l|l|l|l|}
\hline
            & Current & Voltage & Power         \\ \hline
HX711       & 1.5mA   & 5V      & 0.0075W       \\ \hline
RC522       & 13-26mA & 3.3V    & 0.0858W (max) \\ \hline
Arduino Uno & 25.5mA  & 9V      & 0.23W         \\ \hline
Total       &         &         & 320mW         \\ \hline
\end{tabular}
\label{Tab:3}
\end{table}

\section{Future Improvements}
Limited by the development period of this project, the system for now has only the fundamental function to work as a health adherence tool, however, further improvements can be implemented to achieve better performance. \\
\indent The NFC tag can be replaced by a Host-based card emulation in Android, in this way, the MFRC522 will be able to communicate the data with the Phone directly without passing thought the NFC tag media, the data now can be transferred in a more robust and secure manner, however, the battery life will drop as the Smart Pill Case need to be powered after the lid is closed until the data transmission success. \\
\indent An alarm clock function can be added to the software application, the user can input the recommend regimen from the prescription in the setting and the APP will pop up a reminder at the set time.\\
\indent A statistic data analysis can be added to the software application to record and visualize the user's performance on adherence based on the correctness of medication in terms of time and quantities. \\
\indent The shape of the Smart Pill Case can be further improved by fuse the sensors and micro-controllers on a PCB board.\\
\indent As the literature review section indicated, the rate of adherence will increase by increasing the effectiveness of supervision. Cooperate with cloud database solution providers, the data that collected can be further managed and shared with doctors and family members privately to achieve a better adherence performance. 

\section{Conclusion}
In this paper, three existing health adherence tools were reviewed and compared to come up with a new design that can provide real-time feedback of the medicine adherence state on the Smart Phone to the user based on the prescription information input and the sensor data transmitted from the device. The physical part of the device was 3D printed and assembled and the software application was developed in Android Studio. Going through several tests, the device can be conclude that functions properly and fit well with the design criteria. Some potential future improvements were listed for the device to achieve a better performance.

\bibliographystyle{plain}
\bibliography{mybib}

\end{document}